\pdfoutput=1
\documentclass[11pt]{article}

\usepackage[preprint]{acl}

\usepackage{times}
\usepackage{latexsym}
\usepackage[T1]{fontenc}
\usepackage[utf8]{inputenc}
\usepackage{microtype}
\usepackage{graphicx}
\usepackage{booktabs}
\usepackage{xspace}

\usepackage{amsmath}
\usepackage{amssymb}
\usepackage{amsfonts}
\usepackage{bm}
\usepackage{mathtools}
\usepackage{amsthm}
\usepackage{multirow}
\usepackage{wrapfig}
\usepackage{xcolor}
\usepackage{url}

\usepackage{hyperref}
\definecolor{darkbluelinks}{rgb}{0, 0, 0.5}
\hypersetup{colorlinks=true, citecolor=darkbluelinks, linkcolor=darkbluelinks, urlcolor=darkbluelinks}

\usepackage[capitalize,noabbrev]{cleveref}

\theoremstyle{plain}

\theoremstyle{definition}

\theoremstyle{remark}

\newcommand{\rms}{\mathrm{RMS}}

\definecolor{darkblue}{rgb}{0, 0, 0.5}

\definecolor{irenecol}{HTML}{E6194B}   
\definecolor{lorecol}{HTML}{3CB44B}    
\definecolor{valecol}{HTML}{4363D8}    
\definecolor{albecol}{HTML}{F58231}    

\newcommand{\locus}{\textsc{LocUS}}          

\newcommand{\vecs}{\bm{s}} 
\newcommand{\Wu}{\bm{W}_U} 
\newcommand{\rstream}{\bm{r}} 

\DeclareMathOperator{\diag}{diag}
\newcommand{\reals}{\mathbb{R}}
\newcommand{\cS}{\mathcal{S}}
\newcommand{\cP}{\mathcal{P}}

\newcommand{\cV}{\mathcal{V}}
\newcommand{\cA}{\mathcal{A}}

\newcommand{\tet}{\texttt{TET}\xspace}
\newcommand{\imdb}{\texttt{IMDb}\xspace}

\title{\locus{}: Head Selection and Subspace Projection \\
  for Targeted Activation Steering}

\author{
Irene Tallini$^{1}$ \quad Lorenzo Basile$^{1}$\quad Valentino Maiorca$^2$ \\ \textbf{Francesco Locatello}$^{2}$\quad \textbf{Alberto Cazzaniga}$^{1,3}$  \\ [10px]
        $^1$Area Science Park, Trieste, Italy \   $^2$Institute of Science and Technology Austria \\ $^3$Université Côte d’Azur, Inria, LJAD, Maasai Project Team, Nice, France
     \\
     \texttt{\{irene.tallini, alberto.cazzaniga\}@areasciencepark.it}
}
\begin{document}
\maketitle

\begin{abstract}
Activation steering is a powerful training-free paradigm for controlling large language models at inference time. 
However, standard approaches estimate a per-layer steering direction from contrastive data and apply it on the layer's entire representation space, which may couple the intervention to off-target properties present in the contrastive data and degrade unrelated capabilities. To mitigate this issue, we introduce LocUS (Localized Unembedding Steering), a method which grounds activation steering to the model's own output vocabulary subspace. 
By identifying a property-specific linear subspace within the unembedding matrix, LocUS enforces a geometric constraint that restricts the steering transformation to a specific subspace and at the same time localizes its application to a sparse subset of attention heads.
Extensive evaluations across three model families on toxicity mitigation, sentiment redirection and sycophancy suppression show that LocUS matches or outperforms state-of-the-art baselines while intervening on under 6\% of parameters and better preserving general capability.
\end{abstract}

\section{Introduction}
Controlling LLM behavior requires suppressing or amplifying specific properties of generations, e.g. reducing toxicity, shaping tone, or enforcing stylistic constraints. A standard way to achieve this is by fine-tuning, which requires an additional training stage and poses the risk of catastrophic forgetting~\citep{luo2024empirical}. Parameter-efficient variants~\citep{hu2022lora,houlsby2019parameter} reduce training cost but still require training. Prompting provides a lighter alternative, but it is often brittle and difficult to calibrate~\citep{sclar2024quantifying,lu2022fantastically,cheng2026genctrl}. Activation steering offers another option: it modifies the model's internal activations at inference time through transformations estimated from contrastive data~\citep{subramani2022extracting,turner2023activation,zou2023representation,rimsky2024steering,linearact,rodriguez2025end-to-end}.
The additive structure of the residual stream makes vector addition a natural intervention primitive \citep{elhage2021mathematical}, and some properties have been observed to be mediated by a single direction~\citep{arditi2024refusal,li2023inference}.
This view underlies one of the simplest and most widely used steering methods, difference-of-means steering~\citep{turner2023activation,rimsky2024steering}: from two sets of examples that exhibit and lack the target property, one estimates the steering direction as the offset between their mean activations and adds it to the residual stream during the forward pass.

\begin{figure*}[!t]
    \centering
    \includegraphics[width=0.9\textwidth]{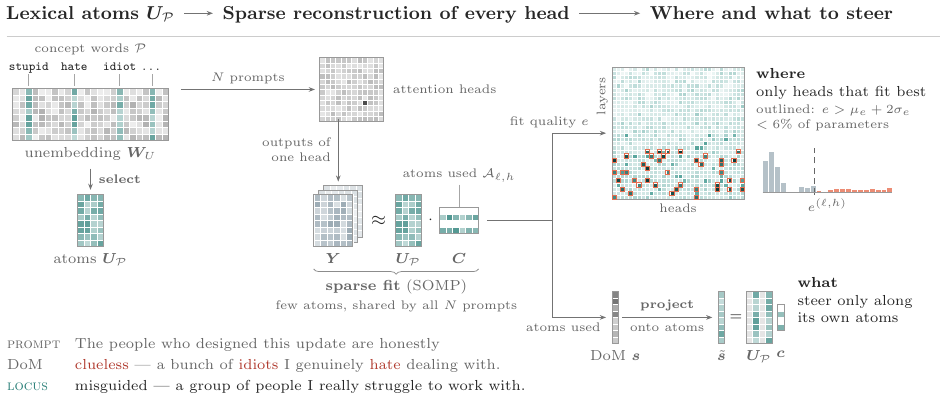}
    \caption{\textbf{The \locus{} pipeline.} A property dictionary $\cP$ (here: \texttt{stupid}, \texttt{hate}, \texttt{idiot}) selects columns of unembedding matrix $\Wu$ to define the atoms matrix $\bm U_\cP$. \emph{(2)} Each head's output is reconstructed through aa subset of atoms in $\bm U_\cP$. Heads with low reconstruction error are retained. The steering vector $\bm s$ is restricted to the per-head selected atoms and added to the retained heads at inference.}
    \label{fig:teaser}
\end{figure*}

The quality of such steering transformation depends on the data used to estimate it: when source and target distributions differ along axes other than the target property, the resulting vector encodes those off-target axes too, and steering along it can damage unrelated capabilities such as factual recall or reasoning~\citep{tan2024analysing}. 
A straightforward mitigation strategy is to localize the intervention to the specific layers and attention heads (residual components) most responsible for the target behavior. But this is not enough, head selection alone leaves the intervention under-constrained: a difference-of-means update applied within the selected heads still carries off-target axes, reintroducing precisely the directions that motivated localization.  

In this paper, we propose \textbf{\locus{}} (\textbf{Loc}alized \textbf{U}nembedding \textbf{S}teering), a fine-grained intervention framework that relies on lexical subspaces to precisely constrain steering. Specifically, \locus{} uses these subspaces to to determine both the \textbf{spatial locus} of the intervention (at attention-head level) and the \textbf{representational subspace} on which the steering transformation should act on.
Both heads and subspaces are isolated by evaluating their alignment with a sparse subset of the columns of the model's unembedding matrix, corresponding to tokens semantically linked to the property of interest/target attribute. These token sets can be automatically generated or user curated, providing a novel interpretability lever for guiding the steering transformation itself. 

To summarize, we propose: 
\begin{enumerate}
    \item \textbf{An automated head selection procedure} that identifies target attention heads by measuring their alignment with lexical content;
    \item \textbf{Subspace-projected steering vectors} that restrict the intervention to a localized, per-head lexical subspace;
    \item \textbf{An empirical evaluation} on three models on toxicity reduction, sentiment control and sycophancy suppression.
\end{enumerate}

\section{Related Work}
While a vast \textit{interpretability} research line has identified property-relevant components at different granularities~\citep{clark2019what,wang2023interpretability,todd2024function,basile2025residual,textspan}, a \textit{steering-oriented} line uses these localized components as intervention sites and evaluates whether the resulting intervention shifts the target metric while preserving unrelated capabilities, at the layer (e.g., ~\citet{rimsky2024steering,wang2025cogsteer}) and, more rarely, head~\citep{li2023inference,izawa2026steering} level.  
\citet{li2023inference} focus of LLaMA thruthfulness steering and select heads by fitting a per-head linear probe; \citet{izawa2026steering} focus on persona traits in two models, first selecting the most influential layer and then, within it, the heads most aligned with the steering vector, making it a method for head selection \textit{inside} a single layer. In both cases, evaluation is tied to a single property, and the number of heads and the intervention strength are either swept or set manually. 
To our knowledge, an automatic, property-agnostic procedure for head-wise steering is thus still lacking in the activation steering literature, and constraining \emph{what} is injected through a property-aligned subspace projection is entirely absent.
\section{Background} 
\label{sec:background}

\paragraph{Difference-of-means steering (DoM).}
Activation steering modifies a model's internal computation at inference time by transforming the activations of one or more intermediate modules, without updating any network weight.

In the vision of \cite{elhage2021mathematical}, transformer language models maintain a \emph{residual stream} that is updated additively at every layer: each attention head and each MLP block produces a contribution that is summed into the stream. Because every module communicates with the rest of the network only through additive writes to this shared channel, injecting an external vector is structurally indistinguishable from adding one more module's output at that layer. This makes vector addition a natural intervention primitive.

One of the simplest and most widely used ways to estimate such a direction is \emph{difference-of-means} (DoM) steering~\citep{turner2023activation, rimsky2024steering}. Given a contrastive dataset of paired examples: a set $\mathcal{D}_+$ that exhibits the target property and a set $\mathcal{D}_-$ that lacks it, one extracts the residual-stream activation $\rstream^{(\ell)}(z)$ for every example $z$ at a chosen intervention point at layer $\ell$, pooling with some function over token positions, and estimates the steering vector as the difference between the class-conditional means:
\begin{equation}\label{eq:dom}
\small
\vecs^{(\ell)} \;=\; \frac{1}{|\mathcal{D}_+|}\!\sum_{z \in \mathcal{D}_+}\!\rstream^{(\ell)}(z) \;-\; \frac{1}{|\mathcal{D}_-|}\!\sum_{z \in \mathcal{D}_-}\!\rstream^{(\ell)}(z).
\end{equation}
At inference, the steered forward pass replaces $\rstream^{(\ell)}$ with $\rstream^{(\ell)} + \alpha\, \vecs^{(\ell)}$, where the scalar $\alpha \in \reals$ controls intervention strength.

DoM has a structural limitation that motivates our work: the estimated direction $\vecs^{(\ell)}$ inherits \emph{every} axis along which $\mathcal{D}_+$ and $\mathcal{D}_-$ differ, not only the target property: any spurious correlation of the labels in the contrastive data contributes to $\vecs^{(\ell)}$ and is injected into the residual stream at steering time~\citep{tan2024analysing}. \locus{} addresses the issue by localizing the intervention to a small set of attention heads which can be used defined and, within each, to a property-aligned subspace.

\paragraph{Head selection with SOMP.}

In \locus{}, attention heads are selected based on by how strongly they align with a subspace defined by tokens relevant for the steering concept. Namely, we adopt the head-scoring procedure of~\citet{basile2025head}, originally developed to identify semantically specialized heads in multimodal transformers. Given a vocabulary subset $\cP \subset \cV$ of concept-relevant tokens (e.g., $\cP = $\{bad, awful, great\} for sentiment steering), the authors define a \emph{property dictionary matrix} by selecting columns of the unembedding matrix $\Wu \in \reals^{d \times |\cV|}$ corresponding to $\cP$:
\begin{equation}
\label{eq:lex_sub}
    \bm U_\cP = \Wu[:,\cP] \in \reals^{d \times |\cP|},
\end{equation}
The columns, called \emph{atoms}, are the residual-stream directions that the model decodes onto the chosen tokens at the final logits.

\begin{table*}[!t]
\centering
\small
\renewcommand{\arraystretch}{1.25}
\begin{tabular}{@{}p{0.46\textwidth}p{0.46\textwidth}@{}}
\toprule
\multicolumn{1}{c}{\textbf{\textsc{Mistral-7B}}} & \multicolumn{1}{c}{\textbf{\textsc{Llama-8B}}} \\
\midrule
\textbf{Prompt:} \textit{POSSIBLE SPOILERS}
&
\textbf{Prompt:} \textit{Early in the movie, Cagney} \\
\addlinespace
\textbf{Steered continuation:} AHEAD. It's been a long time since I watched the first episode of this series. The stupidity and inanities that were present there made me want to throw up. But now, after watching this one, it
&
\textbf{Steered continuation:} 's bad guy is so mean that he even kills a dog. It was such an unnecessary scene and it didn't add anything to the story except for cheap shock value. I don't think I've ever seen any of his movies before unless you count \\
\bottomrule
\end{tabular}
\caption{Qualitative examples of \textsc{LocUS} steering positive$\rightarrow$negative on the IMDb sentiment task.}
\label{tab:imdb_locus_hs_mistral_vs_llama}
\end{table*}

\paragraph{Selecting Heads (\locus{}-h)} \label{par:heads}
\begin{figure}[t]
    \centering
    \includegraphics[width=\columnwidth]{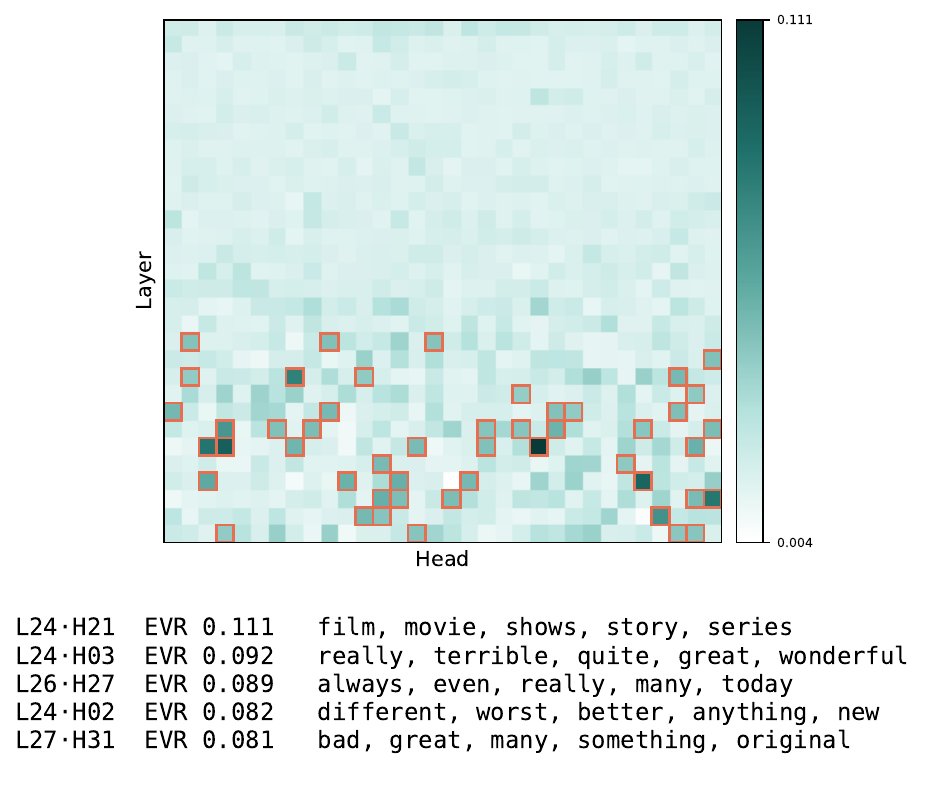}
    \caption{\textbf{Per-head EVR scores and selected atoms.} Above: EVR matrix across all (layer, head) pairs for \imdb{} on deepseek-7b; red squares mark the heads selected by the $\mu+2\sigma$ rule. Below: for the top-5 heads by EVR, the SOMP-selected property atoms in order of selection.}
    \label{fig:score_matrix}
\end{figure}
For each head $(\ell,h)$, one then asks how well its residual contribution $\bm o^{(\ell,h)} \in \reals^d$ can be reconstructed from a sparse combination of property atoms, \emph{consistently across a calibration set} of $N$ examples. Because the unembedding sees activations \emph{after} the final RMSnorm, comparing $\bm o^{(\ell,h)}$ to $\bm U_\cP$ directly would mismatch their respective scales; the authors address this by passing each head output through the final-layer RMSnorm, obtaining $\bar{\bm o}^{(\ell,h)}$.

The reconstruction is then performed by Simultaneous Orthogonal Matching Pursuit (SOMP)~\citep{tropp2006algorithms}, a sparse recovery algorithm that solves a problem of reconstructing a set of vectors with an $n$-dimensional sparse basis drawn from a fixed dictionary matrix by solving it in the form:
\begin{equation}\label{eq:somp}
\min_{\bm C} \;\lVert \bm Y - \bm U_\cP \bm C \rVert_F^2
\quad\text{s.t.}\quad
\lVert \bm C \rVert_{\text{row-0}} \leq n,
\end{equation}
where $\bm Y \in \reals^{d \times N}$ stacks the $N$ calibration activations $\bar{\bm o}^{(\ell,h)}(z)$ as columns, $\bm C \in \reals^{|\cP| \times N}$ holds the per-sample reconstruction coefficients, and $\lVert \cdot \rVert_{\text{row-0}}$ counts the number of nonzero rows of $\bm C$. ``Simultaneous'' refers to the joint support constraint: rather than fitting each sample independently, SOMP forces a single, shared atom set to explain the entire set of samples, so that the selected atoms reflect a stable property of the head. 
The output is a sparse set of tokens $\cA_{\ell,h} \subset \cP$ of size $n$: the property tokens whose unembedding atoms consistently explain the head's output across the calibration set. To measure how well the set of atoms approximates the head output, we consider the \emph{explained-variance ratio} (EVR):
$
e^{(\ell,h)} \;=\; 1 - \frac{\lVert \bm Y - \bm U_\cP \bm C^{(\ell,h)} \rVert_F^2}{\lVert \bm Y \rVert_F^2} \;\in\; [0,1],
$
where $\bm C^{(\ell,h)}$ is the coefficient matrix returned by SOMP with row-support $\cA_{\ell,h}$. A head whose outputs are well captured by a few atoms in $\bm U_\cP$ scores high; a head whose outputs live mostly outside $\mathrm{span}(\bm U_\cP)$ scores low.

\section{Method}
\label{sec:method}

\locus{} uses the idea of lexical subspaces defined in Equation \eqref{eq:lex_sub} to both select heads and, within selected heads, select subspaces where to localize steering. The pipeline is shown in Figure \ref{fig:teaser} and explained, with design choices, below.

In \locus, the SOMP-based head scoring of \cref{sec:background} is repurposed as a steering-site selector, in a fully automatic way.
The property dictionary $\cP$ can be specified manually, generated with an LLM, or derived automatically from the contrastive data; we use the latter choice in our experiments.
We score every head using the procedure in \cref{sec:background} on a calibration set of $N$ concept-bearing examples, with a fixed atom budget. 
Then, we retain the heads whose EVR is unusually large relative to the rest of the network: $
\cS \;=\; \bigl\{(\ell,h) \,:\, e^{(\ell,h)} > \mu_e + 2\sigma_e \bigr\}$, where $\mu_e$ and $\sigma_e$ are the empirical mean and standard deviation of $e^{(\ell,h)}$ across all heads of the model. We denote the resulting data-driven cardinality by $k$ and use it as the number of steering sites for all subsequent operations. The selected heads are depicted with orange squares in \cref{fig:score_matrix} and \cref{fig:score_matrices_full}.

\begin{table*}[!t]
\centering
\small
\setlength{\tabcolsep}{4pt}
\caption{\textbf{\locus{} vs.\ DoM, ITI \citep{li2023inference} and SMH \citep{izawa2026steering}.}
  DoM is the layer-wise difference-of-means update at every layer. Best target and MMLU value per task in
  \textbf{bold}, second best \underline{underlined}.}
  \label{tab:iti_vs_locus}
\resizebox{\textwidth}{!}{%
\begin{tabular}{@{}lccc@{\hspace{1.2em}}ccc@{\hspace{1.2em}}ccc@{}}
\toprule
 & \multicolumn{3}{c}{\textbf{\tet}} & \multicolumn{3}{c}{\textbf{\imdb}} & \multicolumn{3}{c}{\textbf{Sycophancy}} \\
\cmidrule(lr){2-4}\cmidrule(lr){5-7}\cmidrule(lr){8-10}
Method & tox$\downarrow$ (vs U) & MMLU$\uparrow$ & PPL$\downarrow$ & POS$\downarrow$ (vs U) & MMLU$\uparrow$ & PPL$\downarrow$ & SYC$\downarrow$ (vs U) & MMLU$\uparrow$ & PPL$\downarrow$ \\
\midrule
\multicolumn{10}{@{}l}{\emph{Mistral-7B-I}} \\
Unsteered & 0.237 (---) & 0.620 & 7.2 & 0.630 (---) & 0.624 & 6.8 & 66.5 (---) & 0.620 & 7.2 \\
DoM & \underline{0.060} ($-75\%$) & 0.614 & 8.2 & \underline{0.254} ($-60\%$) & 0.620 & 6.9 & 61.5 ($-8\%$) & \textbf{0.619} & 7.4 \\
ITI & 0.102 ($-57\%$) & 0.616 & 8.1 & 0.444 ($-30\%$) & 0.621 & 6.8 & \underline{58.5} ($-12\%$) & 0.612 & 7.6 \\
SMH & 0.075 ($-68\%$) & \textbf{0.623} & 7.6 & 0.552 ($-12\%$) & 0.621 & 6.8 & 68.0 ($+2\%$) & \textbf{0.619} & 7.2 \\
\addlinespace
\locus{} & \textbf{0.020} ($-92\%$) & \underline{0.621} & 8.2 & \textbf{0.009} ($-99\%$) & \textbf{0.624} & 11.6 & \textbf{52.0} ($-22\%$) & \textbf{0.619} & 8.7 \\
\midrule
\multicolumn{10}{@{}l}{\emph{deepseek-7B-I}} \\
Unsteered & 0.241 (---) & 0.514 & 10.7 & 0.658 (---) & 0.483 & 8.3 & 57.0 (---) & 0.514 & 10.7 \\
DoM & 0.042 ($-82\%$) & 0.512 & 12.0 & 0.365 ($-45\%$) & \underline{0.478} & 8.4 & 41.5 ($-27\%$) & \underline{0.513} & 11.4 \\
ITI & 0.332 ($+38\%$) & 0.513 & 11.0 & \underline{0.202} ($-69\%$) & 0.463 & 8.5 & \underline{37.4} ($-34\%$) & 0.512 & 11.4 \\
SMH & \textbf{0.016} ($-93\%$) & 0.509 & 12.8 & 0.450 ($-32\%$) & \underline{0.478} & 8.3 & 53.4 ($-6\%$) & \textbf{0.514} & 10.7 \\
\addlinespace
\locus{} & \underline{0.026} ($-89\%$) & \underline{0.514} & 13.5 & \textbf{0.007} ($-99\%$) & 0.477 & 13.8 & \textbf{14.9} ($-74\%$) & \underline{0.513} & 20.4 \\
\midrule
\multicolumn{10}{@{}l}{\emph{Llama-8B-I}} \\
Unsteered & 0.132 (---) & 0.681 & 9.4 & 0.675 (---) & 0.649 & 8.5 & 80.9 (---) & 0.681 & 9.4 \\
DoM & \underline{0.028} ($-79\%$) & \underline{0.680} & 11.2 & 0.318 ($-53\%$) & \underline{0.647} & 8.6 & 82.5 ($+2\%$) & \textbf{0.681} & 9.5 \\
ITI & 0.145 ($+10\%$) & 0.674 & 9.8 & \underline{0.185} ($-73\%$) & 0.614 & 8.8 & 82.1 ($+1\%$) & \underline{0.678} & 9.8 \\
SMH & 0.115 ($-12\%$) & 0.679 & 9.5 & 0.518 ($-23\%$) & 0.645 & 8.6 & \underline{80.2} ($-1\%$) & \textbf{0.681} & 9.5 \\
\addlinespace
\locus{} & \textbf{0.020} ($-85\%$) & \textbf{0.682} & 14.3 & \textbf{0.119} ($-82\%$) & \textbf{0.649} & 13.6 & 83.8 ($+4\%$) & \underline{0.678} & 11.4 \\
\bottomrule
\end{tabular}%
}
\end{table*}

\paragraph{Selecting Subspaces.}
The dictionary matrix $\bm U_\cP$ suggests another way of constraining steering: instead of using it for head selection, we could use it to directly constrain the steering vector to the property-aligned directions. 
How to do this depends first of all on where in the computational graph we want to act: at layer or head level?

\subparagraph{Layer Level Subspaces (\locus{}-s)}
The residual stream and the property atoms $\bm U_\cP$ live in the same space, so the most direct subspace constraint acts at the residual stream of every layer. We use $\bm U_\cP$ as the dictionary and place an intervention at all layers.

For each layer $\ell$, let $\bar{\bm o}^{(\ell)}\!\in\!\reals^d$ denote the residual-stream response of a calibration prompt at that layer, averaged over tokens and after application of final RMSnorm, for the reason discussed in \cref{sec:background}.
We then run SOMP with dictionary matrix $\bm U_\cP$ and atom budget $n$ to reconstruct the calibration set's $\bar{\bm o}^{(\ell)}$ with a single shared atom set $\cA_\ell$ per layer. Those are the directions in $\bm U_\cP$ that jointly explain the layer's outputs across calibration prompts. The steering vector is thus computed from the reconstructed activations: $\widehat{\bm o}^{(\ell)} \in \mathrm{span}\,\cA_\ell$.
At inference, this reconstructed steering vector is applied to the original activations $\bm o^{(\ell)}$.

\subparagraph{Head Level Subspaces (\locus{}-hs)}
A fair head-level alternative to layerwise steering must intervene with the same total number of steering parameters: a layerwise steering vector lives in $\reals^d$, so intervening on $H$ heads after head output projection would give $H \times d$ parameters and inflate the comparison. We instead intervene on the \emph{raw head output} $\tilde{\bm o}^{(\ell,h)} \in \reals^{d/H}$, i.e., the per-head latent before the output projection.

Constraining the intervention to a sparse property-aligned subspace at this stage requires projecting a raw head activation in the span of the unmebedding directions returned by SOMP.
However, the property dictionary $\bm U_\cP$ lives in residual-stream space, so we need to translate it backwards to raw-head space through the model's decoding path, which results from the composition on three maps: $\bm W_O^{(\ell,h)}$ (attention output matrix, from raw head to residual stream), the final-layer RMSnorm, and the unembedding $\Wu$. Since RMSnorm is non-linear, we linearize it by replacing the per-token reciprocal norm with its calibration-set average over tokens $\hat\rms$, giving the \emph{head-level unembedding matrix}:
\begin{equation}\label{eq:headlogitlens}
\widetilde{\bm U}_\cP^{(\ell,h)} \;=\; \bm U_\cP^\top \,\diag(\gamma \cdot \hat\rms^{-1})\, \bm W_O^{(\ell,h)}
\end{equation}
where $\gamma$ is the final RMSnorm gain. $\widetilde{\bm U}_\cP^{(\ell,h)}$ lives thus in $\reals^{|\cP| \times d/H}$ and each row is the direction in raw-head space whose contribution lands on the corresponding property token at the final logits, akin to a \emph{logit lens} \citep{nostalgebraist2020interpreting} performed at raw head level.

Now, given this matrix, for each head we can compute with SOMP the sparse set of $n$ property-relevant atoms set $\cA_{\ell,h}$.
The steering update is constrained to $\mathrm{span}\,\cA_{\ell,h}$ and applied at inference to the original raw head output $\tilde{\bm o}^{(\ell,h)}$, so each intervention modifies only the components of the head output that the model is expressing along property-aligned directions.

\paragraph{Combining the two: Head Level Subspaces on Selected Heads (\locus{})}
A natural extension of \locus{}-hs is to filter out completely heads we deemed unnecessary with \locus{}-h. Thus, we introduce the complete \locus{} setting that combines head selection with subspace projection: for each $(\ell,h) \in \cS$, the raw-head update is projected onto $\mathrm{span}\,\cA_{\ell,h}$; for heads outside $\cS$, no intervention is applied. The intervention is thus localized along both axes: a sparse set of property-expressing heads, and within each a sparse property-aligned subspace.

\paragraph{Strength normalization.}
Each one of the \locus{} methods changes the norm of the overall
steering update compared to the standard DoM baseline. To compare \locus{} at
matched intervention budget, we rescale each variant so that its total
intervention magnitude heuristically matches that of the unrestricted baseline. Generally
speaking, if we have two modalities of steering $q \in \{q_1, q_2\}$,
intervention sites $i$ with steering vectors at the residual stream level
$\bm s_q^i$, we compute the strength multiplier as
$
\alpha_2 \;=\; \frac{\sum_i \|\bm s_1^i\|_2}{\sum_i \|\bm s_2^i\|_2}.
$
For layerwise variants $\bm s_q^i$ lives in residual-stream space by
construction; for per-head variants the raw-head update is first mapped to the residual stream through the head's output projection. If $\alpha_{1}$ is the
original steering coefficient, $\beta = \alpha_2 \times \alpha_1$. In our case,
$q_1$ is standard DoM and $q_2$ is each \locus{} variant we are testing. $\alpha_1$ is never set by hand: it is the only quantity swept, and the value reported in every table is the one selected on a validation split (\cref{app:strength}).

\begin{table*}[!t]
\centering
\scriptsize
\setlength{\tabcolsep}{2.5pt}
\caption{\textbf{Ablation of the \locus{} components.} \locus{}-h keeps only the EVR head selection ($\mu+2\sigma$, $k$ heads, no subspace projection); \locus{}-s projects the update on the property subspace at the residual stream of every layer (no head selection); \locus{}-hs projects at the raw head output of every head (no selection); \locus{} combines head selection with head-level projection.Best target and MMLU value per task in
  \textbf{bold}, second best \underline{underlined}.}
\label{tab:locus_variants_split50}
\resizebox{\textwidth}{!}{%
\begin{tabular}{@{}lccccc@{\hspace{1.2em}}ccccc@{\hspace{1.2em}}ccccc@{}}
\toprule
       & \multicolumn{5}{c}{\textbf{\tet}} & \multicolumn{5}{c}{\textbf{\imdb}} & \multicolumn{5}{c}{\textbf{Sycophancy}} \\
\cmidrule(lr){2-6}\cmidrule(lr){7-11}\cmidrule(lr){12-16}
Method & $k$ & $\alpha_1$ / $\beta$ & tox$\downarrow$ (vs U) & MMLU$\uparrow$ & PPL$\downarrow$ & $k$ & $\alpha_1$ / $\beta$ & POS$\downarrow$ (vs U) & MMLU$\uparrow$ & PPL$\downarrow$ & $k$ & $\alpha_1$ / $\beta$ & SYC$\downarrow$ (vs U) & MMLU$\uparrow$ & PPL$\downarrow$ \\
\midrule
\multicolumn{15}{@{}l}{\emph{Mistral-7B-I}} \\
Unsteered   & --   & 0    & 0.237 (---)              & 0.620 & 7.2 & --   & 0    & 0.630 (---)              & 0.624 & 6.8 & --   & 0    & 66.5 (---)              & 0.620 & 7.2 \\
\locus{}-h & 42 & 2 / 29.4 & \textbf{0.003} ($-99\%$) & 0.612 & 13.0 & 41 & 4 / 49.5 & 0.012 ($-98\%$) & \underline{0.622} & 11.2 & 42 & 4 / 53.8 & 59.3 ($-11\%$) & 0.615 & 8.3 \\
\locus{}-s & 1024 & 2.5 / 1.10 & \underline{0.010} ($-96\%$) & 0.614 & 10.1 & 1024 & 1.5 / 1.15 & \textbf{0.008} ($-99\%$) & \underline{0.622} & 7.9 & 1024 & 4 / 4.46 & 61.8 ($-7\%$) & \textbf{0.621} & 7.4 \\
\locus{}-hs & 1024 & 3 / 5.15 & 0.036 ($-85\%$) & \underline{0.618} & 8.5 & 1024 & 2 / 3.45 & 0.156 ($-75\%$) & 0.616 & 7.0 & 1024 & 4 / 7.66 & \underline{58.6} ($-12\%$) & \underline{0.620} & 7.4 \\
\addlinespace
\locus{} & 42 & 1 / 24.3 & 0.020 ($-92\%$) & \textbf{0.621} & 8.2 & 41 & 3.5 / 63.7 & \underline{0.009} ($-99\%$) & \textbf{0.624} & 11.6 & 42 & 4 / 105 & \textbf{52.0} ($-22\%$) & 0.619 & 8.7 \\
\midrule
\multicolumn{15}{@{}l}{\emph{deepseek-7B-I}} \\
Unsteered   & --   & 0    & 0.241 (---)              & 0.514 & 10.7 & --   & 0    & 0.658 (---)              & 0.483 & 8.3 & --   & 0    & 57.0 (---)              & 0.514 & 10.7 \\
\locus{}-h & 44 & 0.75 / 7.69 & 0.036 ($-85\%$) & 0.511 & 12.8 & 49 & 2 / 18.2 & \underline{0.013} ($-98\%$) & \textbf{0.482} & 12.9 & 33 & 2 / 29.3 & \underline{31.5} ($-45\%$) & \underline{0.512} & 12.7 \\
\locus{}-s & 960 & 1 / 147 & \textbf{0.020} ($-92\%$) & \underline{0.512} & 14.5 & 960 & 0.75 / 315 & 0.029 ($-96\%$) & \underline{0.478} & 10.6 & 960 & 3 / 1047 & 55.9 ($-2\%$) & \underline{0.512} & 10.9 \\
\locus{}-hs & 960 & 1 / 1.68 & 0.080 ($-67\%$) & 0.511 & 11.4 & 960 & 1 / 1.74 & 0.279 ($-58\%$) & 0.476 & 8.4 & 960 & 2.5 / 4.92 & 47.3 ($-17\%$) & \textbf{0.513} & 11.2 \\
\addlinespace
\locus{} & 44 & 0.75 / 11.7 & \underline{0.026} ($-89\%$) & \textbf{0.514} & 13.5 & 49 & 2 / 28.9 & \textbf{0.007} ($-99\%$) & 0.477 & 13.8 & 33 & 3.5 / 95.7 & \textbf{14.9} ($-74\%$) & \textbf{0.513} & 20.4 \\
\midrule
\multicolumn{15}{@{}l}{\emph{Llama-8B-I}} \\
Unsteered   & --   & 0    & 0.132 (---)              & 0.681 & 9.4 & --   & 0    & 0.675 (---)              & 0.649 & 8.5 & --   & 0    & 80.9 (---)              & 0.681 & 9.4 \\
\locus{}-h & 36 & 0.75 / 11.4 & \textbf{0.016} ($-88\%$) & 0.674 & 12.1 & 44 & 2 / 39.2 & \textbf{0.070} ($-90\%$) & \textbf{0.652} & 16.1 & 32 & 3 / 74.4 & 82.2 ($+2\%$) & \underline{0.680} & 11.0 \\
\locus{}-s & 1024 & 4 / 9.20 & 0.024 ($-81\%$) & \textbf{0.683} & 18.7 & 1024 & 3.5 / 12.9 & 0.264 ($-61\%$) & 0.641 & 10.5 & 1024 & 0.5 / 2.92 & \textbf{81.7} ($+1\%$) & \textbf{0.681} & 9.4 \\
\locus{}-hs & 1024 & 2 / 3.58 & 0.037 ($-72\%$) & 0.676 & 10.2 & 1024 & 1.5 / 2.72 & 0.369 ($-45\%$) & 0.648 & 8.7 & 1024 & 2 / 3.92 & \underline{82.1} ($+1\%$) & \textbf{0.681} & 9.5 \\
\addlinespace
\locus{} & 36 & 0.75 / 19.2 & \underline{0.020} ($-85\%$) & \underline{0.682} & 14.3 & 44 & 1.5 / 50.5 & \underline{0.119} ($-82\%$) & \underline{0.649} & 13.6 & 32 & 3 / 159 & 83.8 ($+4\%$) & 0.678 & 11.4 \\
\bottomrule
\end{tabular}%
}
\end{table*}

\section{Experimental Setup}
\label{sec:experiments}
We evaluate \locus{} on three tasks: toxicity mitigation on ThoroughlyEngineeredToxicity~\citep{luong2024realistic} (\tet, QA format), sentiment redirection on \imdb~\citep{maas2011learning} (free-form continuation, positive$\rightarrow$negative) and sycophancy suppression on the persona data of \citet{chen2025persona} (QA format). We evaluate three model families: Mistral-7B~\citep{jiang2023mistral}, deepseek-7b~\citep{bi2024deepseek}, and Llama-8B~\citep{dubey2024llama}, using the Instruct/Chat variants on \tet{} and Sycophancy (matching their QA format) and the base variants on \imdb{} (which calls for free-form continuation).

\paragraph{Baselines and metrics.}

We compare \locus{}-h and \locus{} against the unsteered model and against standard DoM. As ablations for the subspace projection variant, we report \locus{}-s, \locus{}-hs. We further compare \locus{} with the two existing head-level methods: ITI \citep{li2023inference}, which selects the same number $k$ of heads through linear probes, and SMH \citep{izawa2026steering}, which selects a single layer and the heads within it most aligned with the steering vector; both use their published strength grids. The target metric is the toxicity rate for \tet (as measured using the RoBERTa classifier of~\citet{logacheva2022paradetox}), the positive rate for \imdb (measured with DistilBERT SST-2 \citep{sanh2019distilbert}) and, for Sycophancy, the 0--100 trait score assigned by an LLM judge following \citet{chen2025persona}. In all cases, lower values represent more effective steering. To measure the impact of steering over general capabilities unrelated to the steering target, we use MMLU~\citep{hendrycks2021measuring}, a 57-subject 5-shot multiple-choice test that probes factual knowledge through closed answers. We report perplexity (PPL) on the openings of 20k Wikipedia articles, following the protocol of \citet{rodriguez2025end-to-end} (disjoint 10k validation and test halves; \cref{app:impl_eval}), as a sanity check that generation is not gibberish; we treat a PPL above twice that of the unsteered model as degraded output.

\paragraph{Hyperparameters.}
\locus{} has two free hyperparameters: the SOMP atom budget at head selection and at subspace projection. We fix the former to 50 atoms as in \citet{basile2025head} and set the latter to $n{=}20$ from the fit data alone (\cref{app:granularity}). The number of selected heads and the multiplier $\alpha_2$ are determined automatically from calibration data, while the coefficient $\alpha_1$ is swept on a validation split and fixed at the value that minimizes the target metric among those whose validation wiki-PPL is at most twice the unsteered model's and whose validation MMLU is at least $0.99\times$ the unsteered value (\cref{app:strength}); the test numbers in all tables are obtained at that single value. Full evaluation protocol is reported in \cref{app:impl}.

\section{Experimental Results}
\label{sec:results}

In the majority of cases, selecting heads beats steering on whole layers, and selecting subspaces within heads beats selecting heads alone. Moreover, as an overall steering method, \locus{} proves more effective than existing head-level methods \citep{li2023inference,izawa2026steering} in most settings. We will go into details below. 
In \cref{tab:imdb_locus_hs_mistral_vs_llama} we show two example generations. \textcolor{red}{Note: some of the generations can be offensive.}

\paragraph{\locus{} vs.\ layerwise and head-level baselines (\cref{tab:iti_vs_locus}).}
At the validation-selected strength, \locus{} reaches a lower target than DoM in eight of the nine cells, the ninth being Sycophancy on Llama-8B, where, however, no steering method moves the score, an thus we consider it a failure case for difference-of-means steering in general. It is also lower than ITI in the same eight cells. The reason we identified is that a probe ranks heads by how well they \emph{detect} the property whereas EVR ranks them by how strongly they \emph{write} it: \Cref{fig:sorted-probes} shows the property is linearly decodable from a broad set of heads while \Cref{fig:sorted-evr} shows only a few write strongly onto it. SMH, confined to a single layer and to its published strengths, understeers everywhere except on \tet{}/deepseek, the only cell where it beats \locus{}. The sweep in \cref{fig:pareto-val} shows what \locus{} pays for this: DoM loses MMLU as the strength grows while \locus{} keeps it at the unsteered value and loses fluency instead, which is why its PPL is the highest in most cells.

\paragraph{\locus{} variants ablations (\cref{tab:locus_variants_split50} and \cref{fig:pareto-val}).}
Head selection alone (\locus{}-h) is below DoM in every cell and accounts for most of the gain. Projection on top of it protects capabilities further. Projection without selection does not help: \locus{}-hs spreads the update over all heads and loses MMLU like DoM, while \locus{}-s keeps MMLU but understeers. The one cell where \locus{} trails \locus{}-h (\imdb{}/Llama-8B, $0.119$ vs.\ $0.070$) is a PPL effect: concentrating the budget on fewer directions makes PPL grow faster, and the next strength on the grid is rejected by the gate.

\begin{table}[!t]
\centering
\small
\setlength{\tabcolsep}{4pt}
\caption{\textbf{Steering fitted on an external corpus (Jigsaw), evaluated on \tet{}.}
  Best target and MMLU value per task in
  \textbf{bold}.}
  \label{tab:jigsaw_fit}
\resizebox{\columnwidth}{!}{%
\begin{tabular}{@{}lccccc@{}}
\toprule
Method & $k$ & $\alpha_1$ / $\beta$ & tox$\downarrow$ (vs U) & MMLU$\uparrow$ & PPL$\downarrow$ \\
\midrule
\multicolumn{6}{@{}l}{\emph{Mistral-7B-I}} \\
Unsteered & -- & 0 & 0.237 (---) & 0.618 & 7.2 \\
DoM & -- & 4 / 4 & 0.0943 ($-60\%$) & 0.618 & 8.7 \\
\locus{} & 19 & 0.5 / 27.2 & \textbf{0.0358 ($-85\%$)} & \textbf{0.619} & 8.3 \\
\midrule
\multicolumn{6}{@{}l}{\emph{Llama-8B-I}} \\
Unsteered & -- & 0 & 0.132 (---) & 0.683 & 9.4 \\
DoM & -- & 4 / 4 & 0.0374 ($-72\%$) & 0.677 & 11.9 \\
\locus{} & 19 & 0.5 / 20.9 & \textbf{0.0325 ($-75\%$)} & \textbf{0.679} & 13.4 \\
\midrule
\multicolumn{6}{@{}l}{\emph{deepseek-7B-I}} \\
Unsteered & -- & 0 & 0.241 (---) & 0.513 & 10.7 \\
DoM & -- & 3.5 / 3.5 & 0.0260 ($-89\%$) & \textbf{0.513} & 13.3 \\
\locus{} & 32 & 1 / 19.7 & \textbf{0.0033 ($-99\%$)} & 0.509 & 14.9 \\
\bottomrule
\end{tabular}%
}
\end{table}

\paragraph{Fitting toxicity on an external corpus (\cref{tab:jigsaw_fit}).}
With dictionary, head scores, subspaces and steering vector all fitted on human-written Jigsaw comments, \locus{} still reduces toxicity more than DoM fitted on the same data on all three models ($85\%$, $75\%$ and $99\%$ vs.\ $60\%$, $72\%$ and $89\%$), with MMLU within $0.005$ of the unsteered model.

\begin{table*}[!t]
\centering
\scriptsize
\setlength{\tabcolsep}{2.5pt}
\caption{\textbf{DoM vs.\ \locus{} on deepseek-67B, test split.} Best target and MMLU value per task in
  \textbf{bold}.}
  \label{tab:ds67b_test}
\resizebox{\textwidth}{!}{%
\begin{tabular}{@{}lccccc@{\hspace{1.2em}}ccccc@{\hspace{1.2em}}ccccc@{}}
\toprule
 & \multicolumn{5}{c}{\textbf{\tet}} & \multicolumn{5}{c}{\textbf{\imdb}} & \multicolumn{5}{c}{\textbf{Sycophancy}} \\
\cmidrule(lr){2-6}\cmidrule(lr){7-11}\cmidrule(lr){12-16}
Method & $k$ & $\alpha_1$ / $\beta$ & tox$\downarrow$ (vs U) & MMLU$\uparrow$ & PPL$\downarrow$ & $k$ & $\alpha_1$ / $\beta$ & POS$\downarrow$ (vs U) & MMLU$\uparrow$ & PPL$\downarrow$ & $k$ & $\alpha_1$ / $\beta$ & SYC$\downarrow$ (vs U) & MMLU$\uparrow$ & PPL$\downarrow$ \\
\midrule
Unsteered & -- & 0 & 0.286 (---) & 0.721 & 5.8 & -- & 0 & 0.652 (---) & 0.714 & 5.8 & -- & 0 & 75.8 (---) & 0.721 & 5.8 \\
DoM & -- & 2 / 2 & 0.020 ($-93\%$) & 0.714 & 7.3 & -- & 1 / 1 & 0.387 ($-41\%$) & 0.709 & 5.9 & -- & 4 / 4 & 58.2 ($-23\%$) & \textbf{0.721} & 6.0 \\
\locus{} & 166 & 1 / 18.2 & \textbf{0.000} ($-100\%$) & \textbf{0.728} & 9.1 & 186 & 2 / 28.1 & \textbf{0.009} ($-99\%$) & \textbf{0.713} & 9.2 & 173 & 4 / 106 & \textbf{54.4} ($-28\%$) & 0.714 & 9.4 \\
\bottomrule
\end{tabular}%
}
\end{table*}

\paragraph{Scaling to 67B (\cref{tab:ds67b_test}).}
The picture is unchanged at 67B: \locus{} selects about $3\%$ of the $6{,}080$ heads and beats DoM on all three tasks ($0.000$ vs.\ $0.020$ on \tet{}, $0.009$ vs.\ $0.387$ on \imdb{}, $54.4$ vs.\ $58.2$ on Sycophancy), with MMLU within $0.01$ of the unsteered model and the same trade-off as at 7B.

\paragraph{Parameter count and degrees of freedom.}
The full \locus{} pipeline intervenes on at most $5.1\%$ of attention heads, and within each retained head confines the steering update to a per-head subspace of $16\%$ of the per-head dimensionality. Multiplying through, the total number of degrees of freedom used by \locus{} is between $0.4\%$ and $0.8\%$ of the $L \cdot d$ degrees of freedom of an unprojected DoM update applied at every layer. The improvements reported above are therefore achieved through a substantially smaller, structurally and geometrically localized intervention.

\section{Conclusions}
\label{sec:conclusions}
We presented \locus{}, a targeted steering procedure that confines difference-of-means activation steering to a small set of attention heads and, within each, to a sparse subspace of property-aligned directions selected through a user-controllable lexical dictionary. Across three model families and three tasks, this two-level localization yields stronger target-metric shifts than layerwise difference-of-means and existing head-level steering methods \citep{li2023inference,izawa2026steering}, while preserving the general capabilities computed through MMLU, using a fraction of the steering parameters of unrestricted interventions.
\paragraph{Future Work.}
MMLU covers knowledge retrieval on a wide variety of topics but does not evaluate fine-grained controllability (e.g.\ suppressing toxicity while preserving 
specific demographic content), which requires a dedicated benchmark.
Contrary to prior head-selection criteria, which are derived from contrastive or probing data and therefore propagate whatever bias that data carries into the heads selection procedure itself, \locus{} allows interpretable filtering of the lexical directions we care most about (e.g., when steering for toxicity reduction, removing from the dictionary atoms related to demographic content). Thus it would be interesting to analyze fine-grained controllability also at different curation levels of the property dictionary.
Because \locus{} only specifies \emph{where} and \emph{in what subspace} to intervene, it is in principle compatible with any steering method. Composing \locus{} with more sophisticated estimators (see e.g. \cite{rodriguez2025end-to-end}) is a natural direction for future work.
\paragraph{Limitations}
Main limitation of this method is that, as it is, it is only applicable in settings where the property is carried by some lexical difference. The method also tends to degrade fluency (as computed by PPL) more than factual reasoning. Capability is measured with MMLU and PPL alone, so effects on open-ended reasoning are not covered.

\section*{Acknowledgments}
\looseness=-1The authors acknowledge the Area Science Park supercomputing platform ORFEO made available for conducting the research reported in this paper, and the technical support of the Laboratory of Data Engineering staff.
IT and AC were supported by the project ``Supporto alla diagnosi di malattie rare tramite l’intelligenza artificiale" CUP: F53C22001770002 and ``Valutazione automatica delle immagini diagnostiche tramite l’intelligenza artificiale", CUP: F53C22001780002. LB was supported by the European Union – NextGenerationEU within the project PNRR ``Finanziamento di progetti presentati da giovani ricercatori" - Mission 4 Component 2 Investment 1.2, CUP: J93C25000440001. VM was supported by the ISTA Responsible AI Program, made possible through the generous support of Garrett Camp and the Camp Foundation. FL was funded in part by the Austrian Science Fund (FWF) 10.55776/COE12. AC was supported by the European Union – NextGenerationEU within the project PNRR ``PRP@CERIC" IR0000028 - Mission 4 Component 2 Investment 3.1 Action 3.1.1. 

\bibliography{references}

\clearpage
\appendix

\section*{Ethical Considerations}
Our objective is to contribute to safer and more controllable language models. Methods that enable targeted modification of model behavior can support interpretability, auditing, and alignment; however, the same capabilities could be misused to suppress, emphasize, or disguise particular types of content. We therefore urge users and practitioners to assess potential downstream uses carefully, particularly when applying these methods in high-stakes or sensitive settings. The models and datasets used in this work comply with the terms of their respective research licenses. Upon acceptance, we will make the code available for non-commercial research use under compatible licensing conditions. AI-assisted programming tools, including Claude Code, were used only to facilitate code completion during implementation. All AI-generated code was inspected, validated, and overseen by the authors.

\section{Implementation details}
\label{app:impl}

\subsection{Models.}
Our IMDb (sentiment) experiments use the base checkpoints mistralai/Mistral-7B-v0.1, meta-llama/Llama-3.1-8B, and
  deepseek-ai/deepseek-llm-7b-base, while the TET (toxicity) and Sycophancy experiments use the instruction-tuned counterparts
  mistralai/Mistral-7B-Instruct-v0.3, meta-llama/Llama-3.1-8B-Instruct, and deepseek-ai/deepseek-llm-7b-chat. All models are obtained from their
  official Hugging Face repositories and used at their released revisions in bfloat16, with no further fine-tuning.
\subsection{Datasets and pipeline}
\label{app:datasets}

\paragraph{\tet{} (toxicity).}
The 2{,}546 prompts of ThoroughlyEngineeredToxicity~\citep{luong2024realistic} are split in half: 1{,}273 are held out for test and the remaining ones are split again into 636 fit and 637 validation prompts. For each fit prompt we sample completions from the unsteered model and label them with the RoBERTa toxicity classifier of~\citet{logacheva2022paradetox}. The toxic completions (up to $512$) and an equal number of non-toxic ones form the contrastive fit set for the steering vector, the property dictionary and the EVR scores.

\paragraph{\imdb{} (sentiment).}
The fit set is $512$ positive and $512$ negative reviews from the IMDb train split. Evaluation prompts are the first $8$ tokens of a test-split review, kept only if they contain no word of the Hu--Liu sentiment lexicon; the first $1{,}000$ such prefixes form the test set and the next $1{,}000$ the validation set.

\paragraph{Sycophancy.}
We use the sycophancy trait data of \citet{chen2025persona}: $512$ trait-eliciting and $512$ neutral responses, drawn from disjoint prompt halves of the train split, form the fit set. Evaluation follows the same work: $5$ trait-eliciting system prompts $\times$ $20$ questions, scored $0$--$100$ by Qwen2.5-32B-Instruct with the original judge prompt; the extraction questions serve as validation and the evaluation questions as test.

\paragraph{Fit/validation/test usage.}
For all tasks the \emph{fit} split is used for (i) the property dictionary (\cref{eq:klwords}), (ii) the per-head EVR matrix, (iii) the per-head dictionaries $\bm D^{(\ell,h)}$ and (iv) the difference-of-means steering vector. The \emph{validation} split selects the steering coefficient $\alpha_1$ under the capability gate, and the \emph{test} split is used only for the metrics reported in the tables.

\paragraph{Computational resources}
We ran the experiments on one NVIDIA H100
GPU. We used the HuggingFace Transformers
library \citep{wolf2020transformers} to obtain publicly available checkpoints for the Mistral, Llama and Deepseek models. The total compute time for each experiment is approximately 0.5 GPU hours.

\subsection{Method}
\label{app:impl_method}
\paragraph{Property dictionary.}
Let $p_S(w)$ and $p_D(w)$ denote the Dirichlet-smoothed ($\alpha{=}0.01$) unigram probabilities of word $w$ under the property-bearing and non-property-bearing fit corpora. We score each word by its pointwise contribution to the KL divergence,
\begin{equation}\label{eq:klwords}
\mathrm{KL}_w \;=\; p_S(w) \log \frac{p_S(w)}{p_D(w)},
\end{equation}
in both directions, remove stop words, keep the $70$ highest-scoring words of each direction and take their union, dropping non-ASCII entries; this yields $139$--$140$ words per (model, task) cell, which we tokenize to obtain $\cP$ and the property subspace $\bm U_\cP = \Wu[:, \cP]$. On roughly 1/3 of the cells \tet{} the automatic list is further manually filtered to remove too generic terms.

\paragraph{Steering setting.}
Steering vectors are estimated using the \emph{atonce} strategy: for every selected head, we run a single forward pass over the calibration set with all hooks active and capture activations in parallel, then estimate each head's difference-of-means update on those activations. This contrasts with an \emph{incremental} strategy, in which one would estimate the update for the first layer's hook, apply it, then estimate the next layer's update on the already-steered activations, and so on. The atonce variant is faster and avoids hook-order dependence at the cost of ignoring downstream effects of upstream interventions.

The steering vector is computed from activations \emph{averaged over the sequence positions} of the prompt. At inference, the resulting steering vector is applied only to the activation at the \emph{last} token position of the generated continuation.

\subsection{Evaluation}
\label{app:impl_eval}

\paragraph{Steering coefficient selection.}
\label{app:strength}
The coefficient $\alpha_1$ is the only swept quantity, and it is chosen once per (method, model, task) cell on the validation split; the tables report the test run at the selected value $\alpha_1^*$. For all locus variants and DoM the sweep grid is: $\alpha_1\in\{0.25,0.5,0.75,1,1.5,\dots,4\}$. Reference methods are swept with the grid indicated in the respective apper $\alpha\in\{5,10,\dots,30\}$ for ITI, and $\alpha\in\{0.5,\dots,14\}$ for SMH. A grid point is admissible if its validation wiki-PPL is at most twice the unsteered model's and its validation MMLU is at least $0.99\times$ the unsteered value; $\alpha_1^*$ is the admissible point with the lowest validation target metric.

\paragraph{Generation.}
On \tet{} we sample $2$ completions of $64$ tokens for each of the $615$ test prompts ($308$ on validation); on \imdb{} one completion of $50$ tokens for each of the $1{,}000$ prompts; on Sycophancy $2$ completions of $64$ tokens for each of the $100$ (system prompt, question) pairs. Sampling uses temperature $1.0$, with top-$p$ $0.3$ and repetition penalty $1.2$ on \tet{} and \imdb{}.

\paragraph{Control Metrics.}
\texttt{PPL-Wiki} follows the protocol of \citet{rodriguez2025end-to-end}: it is the mean per-passage perplexity over 20k English Wikipedia articles (Wikimedia 20231101 dump), each scored from its opening on the first 50 tokens; validation and test use disjoint halves (10k passages each). The degeneration gate is relative: an operating point is admissible if its validation PPL-Wiki is at most twice that of the unsteered model. \texttt{MMLU} is the 57-subject 5-shot accuracy computed with the EleutherAI \texttt{lm-evaluation-harness} \citep{eval-harness}; its 14{,}042 items are split once, stratified by subject, into a validation half used by the selection gate and a test half reported in the tables (the Jigsaw and 67B tables use the full set).

\paragraph{ITI baseline.} 
Following \citep{li2023inference}., we train one logistic-regression probe per attention head (sklearn LogisticRegression, C=1.0, L2, max$\_$iter=1000, no feature standardization) on the last-token per-head activations (the o$\_$proj inputs), with classes balanced to the minority count and 2-fold cross-validation, and rank heads by mean held-out accuracy, selecting the top K with K set to LOCuS-h's head count.

\paragraph{Where the budget $n{=}20$ comes from.}\label{app:granularity}

\begin{wraptable}{r}{0.5\columnwidth}
\vspace{-1.2em}
\centering
\caption{\textbf{Choosing the subspace budget from the effective dictionary.} Smallest number of atoms at which the mean
fit-time cumulative EVR over all heads exceeds $0.25$.}
\label{tab:natoms_evr}
\scriptsize
\setlength{\tabcolsep}{5pt}
\begin{tabular}{@{}llc@{}}
\toprule
Task & Model & $n^\star$ \\
\midrule
\tet{}       & Mistral-7B  & 22 \\
             & deepseek-7B & 20 \\
             & Llama-8B    & 24 \\
             & deepseek-67B & 17 \\
\addlinespace
\imdb{}      & Mistral-7B  & 24 \\
             & deepseek-7B & 22 \\
             & Llama-8B    & 25 \\
             & deepseek-67B & 22 \\
\addlinespace
Sycophancy   & Mistral-7B  & 19 \\
             & deepseek-7B & 16 \\
             & Llama-8B    & 19 \\
             & deepseek-67B & 14 \\
\midrule
\multicolumn{2}{@{}l}{median} & 21 \\
\bottomrule
\end{tabular}
\end{wraptable}
The value used in the main tables is fixed from the fit data alone, with no access to any
steering metric. For each
(task, model) cell we replay the fit-time SOMP decomposition of \cref{eq:somp} on the
effective per-head dictionaries $\widetilde{\bm U}_\cP^{(\ell,h)}$ with a budget of $50$
atoms, which yields, for every head, the cumulative EVR $e^{(\ell,h)}(n)$ of \cref{sec:background} as a function of
the number of atoms retained. We then read off $
n^\star \;=\; \min\Bigl\{\, n \;:\; \operatorname*{mean}_{(\ell,h)} \mathrm{EVR}^{(\ell,h)}(n) \;>\; 0.25 \,\Bigr\},$
the smallest budget at which the average head has a quarter of its fit-time variance
reconstructed by its own atoms, with the average taken over all heads. \Cref{tab:natoms_evr} reports $n^\star$:
over the twelve cells it lies in $[14,25]$ (median $21$), and no cell asks for a budget far
from $20$, which motivates the single value $n{=}20$ used for \locus{} and \locus{}-hs
throughout \cref{tab:locus_variants_split50}. The rule is defined for the per-head
dictionaries; for the per-layer \locus{}-s we set $n{=}10$ by hand.

\section{Head selection matrices}
\label{app:headsel}
\begin{figure*}[t]
\centering
\includegraphics[width=0.8\textwidth]{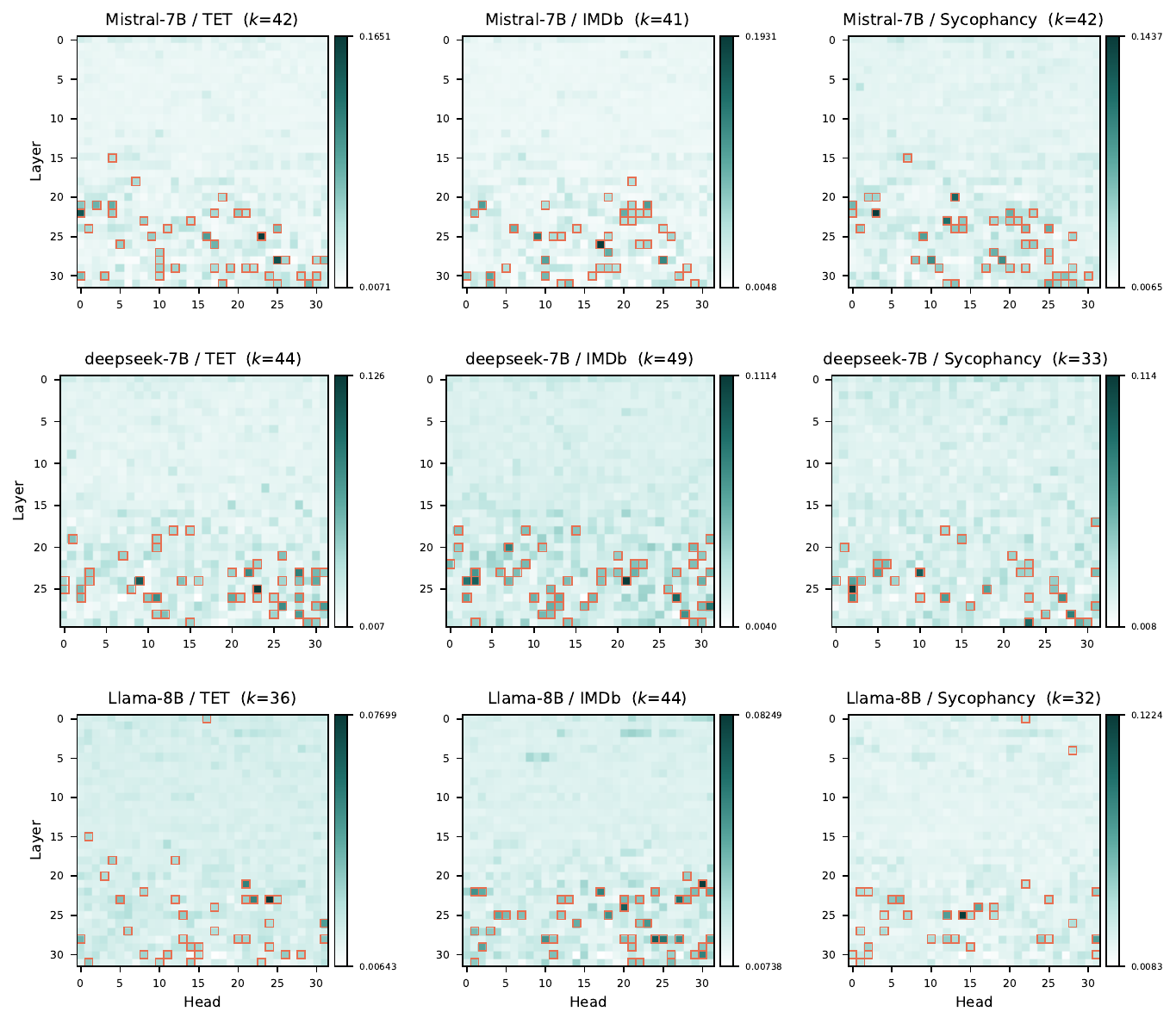}
\caption{Per-head EVR matrices for the nine (model, task) cells behind \cref{tab:locus_variants_split50}; orange boxes mark the $\mu+2\sigma$-selected heads and $k$ is their count.}
\label{fig:score_matrices_full}
\end{figure*}

\begin{figure*}[t]
    \centering
    \includegraphics[width=0.8\textwidth]{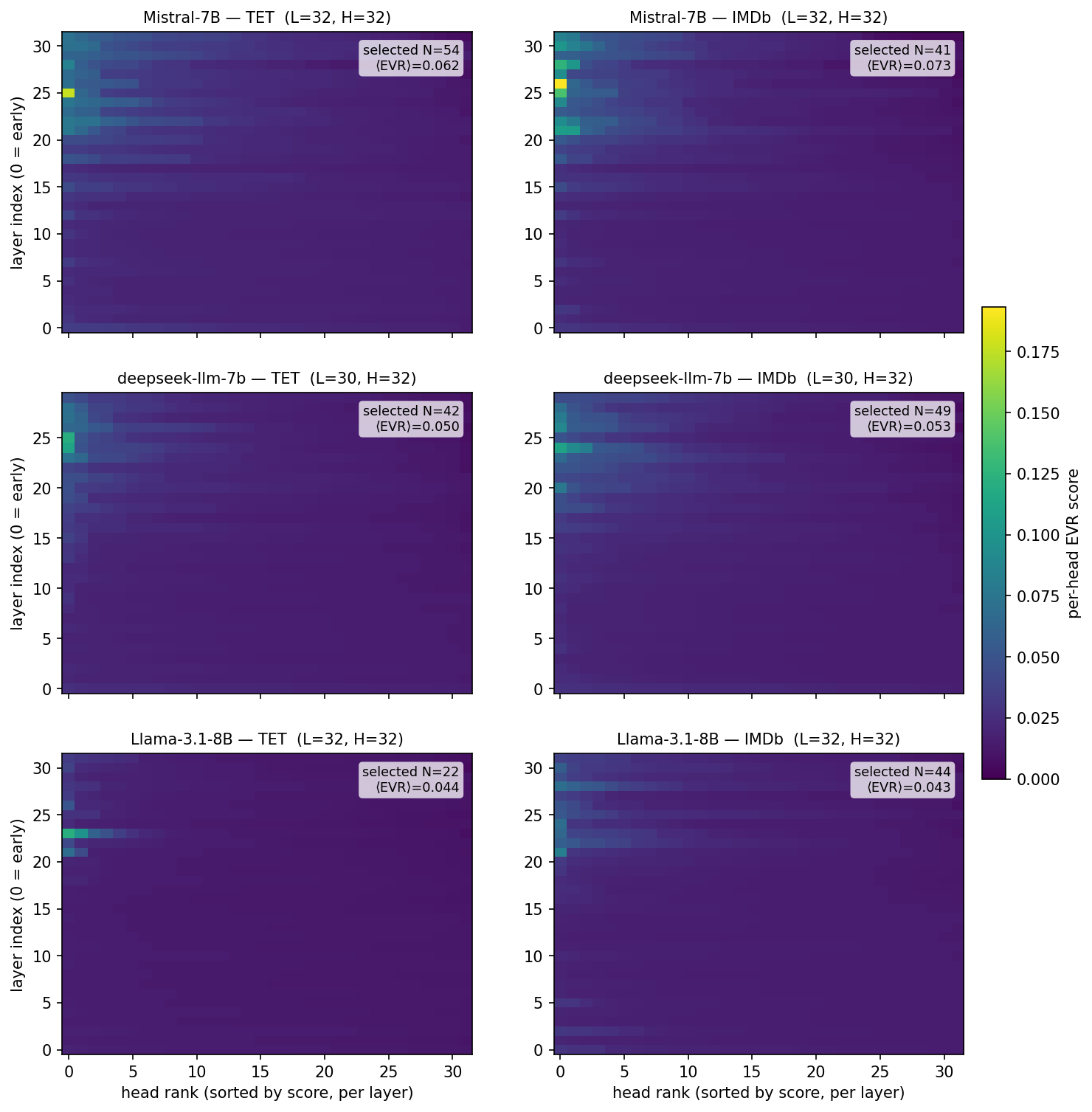}
    \caption{EVR on heads sorted by EVR value. Per-head EVR scores for the nine (model, task) cells, heads sorted by EVR
within each layer (rank $0$ = highest). Inset reports the number of heads selected
by the $\mu+2\sigma$ rule and their mean EVR.}
    \label{fig:sorted-evr}
\end{figure*}

\begin{figure*}[t]
    \centering
    \includegraphics[width=0.8\textwidth]{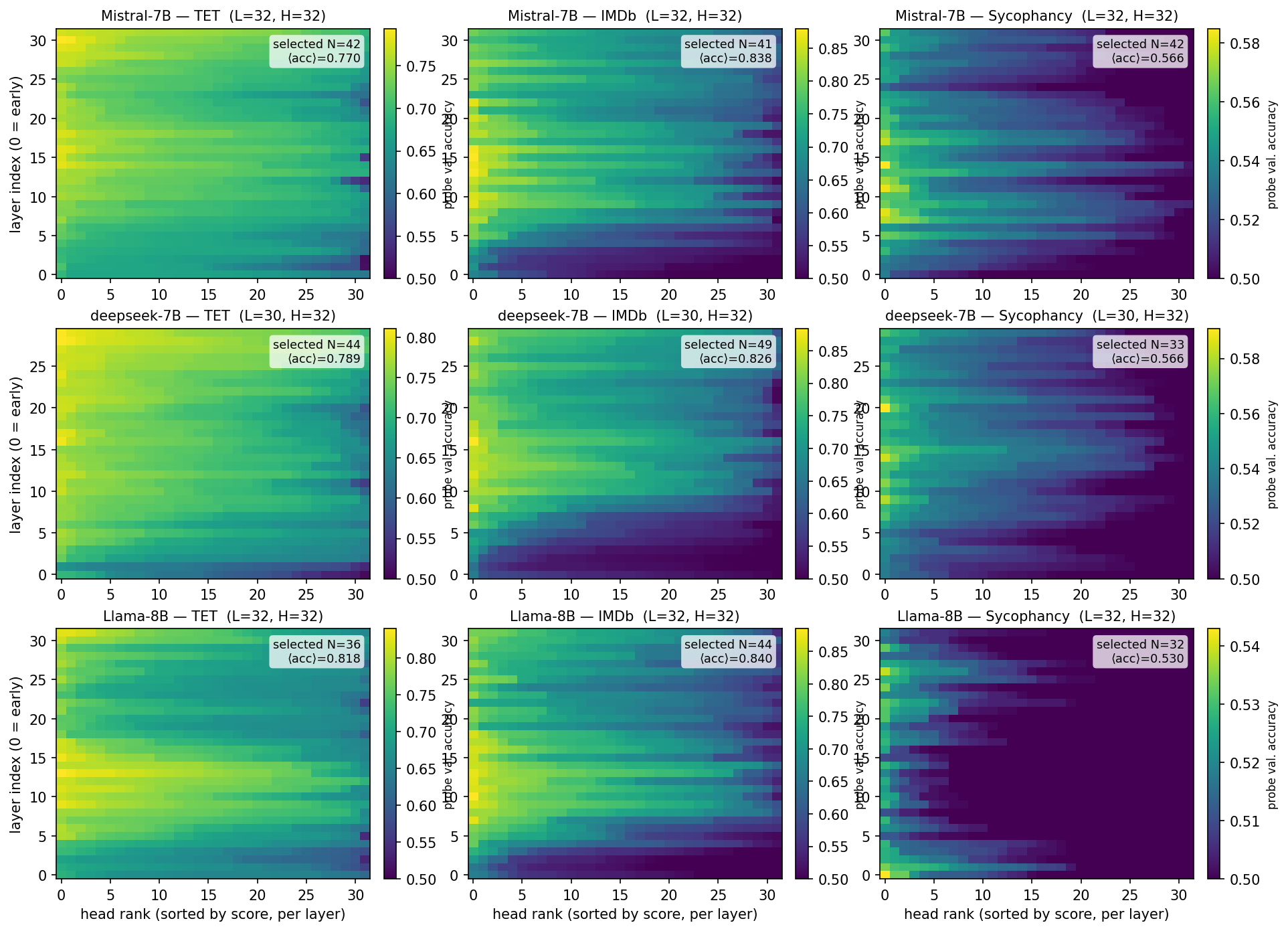}
    \caption{Sorted probe accuracy for ITI. Per-head linear-probe validation accuracy for the nine (model, task)
cells, same layout as Figure~\ref{fig:sorted-evr}. Inset reports the head count
selected by ITI (the same $k$ as \locus{}) and their mean probe accuracy.}
    \label{fig:sorted-probes}
\end{figure*}

Figures~\ref{fig:score_matrices_full}, \ref{fig:sorted-evr}, and~\ref{fig:sorted-probes} visualize the per-head signals underlying \locus{} and the ITI baseline.
Figure~\ref{fig:score_matrices_full} plots the EVR matrix over all (layer, head) pairs,
with heads retained by the $\mu+2\sigma$ rule marked in orange;
Figure~\ref{fig:sorted-evr} shows the same scores with heads ranked by EVR within
each layer.

Figure~\ref{fig:sorted-probes} shows the linear-probe validation accuracy used by
ITI, in the same layout. Unlike EVR, high probe accuracy is broadly distributed
across heads and layers.
\label{app:score_matrices_full}

\section{Strength Comparison}
\label{app:strength_comp}

\begin{figure*}
    \centering
    \includegraphics[width=\textwidth]{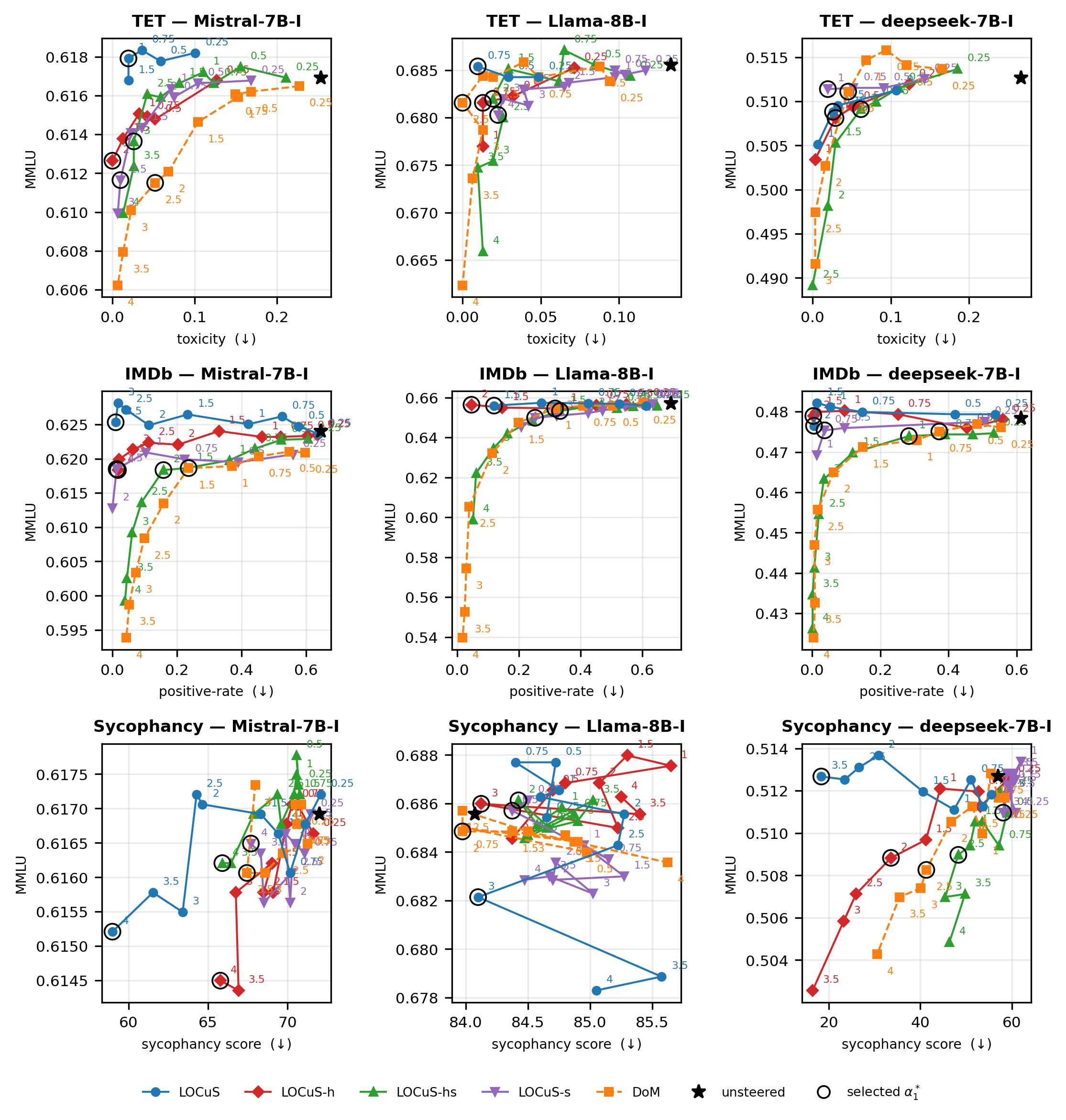}
    \caption{\textbf{Removal--capability trade-off over the strength sweep, validation split.}
    Each panel sweeps the steering coefficient $\alpha_1\in\{0.25,\dots,4\}$ (annotated next to
    each marker) for one (task, model) cell and plots the validation target metric ($x$,
    lower is better) against validation MMLU ($y$, higher is better); the black star is the unsteered model. Points whose validation wiki-PPL exceeds twice the
    unsteered model's are omitted (MMLU itself is not gated in the plot); A black ring marks each variant's selected $\alpha_1^*$, i.e.\ the
    operating point \cref{tab:locus_variants_split50} reports at test.}
    \label{fig:pareto-val}
\end{figure*}

\Cref{fig:pareto-val} shows the full validation sweep behind \cref{tab:iti_vs_locus,tab:locus_variants_split50}: for every (task, model) cell, the target metric against MMLU as $\alpha_1$ grows, for DoM and the four \locus{} variants, with the selected $\alpha_1^*$ circled. On Sycophancy with Llama-8B no variant, at any strength, moves the score away from the unsteered value (the whole panel spans less than two points of sycophancy score), which is why every method reports an unchanged target on that cell.

\end{document}